# Magnetic Milli-spinner for Robotic Endovascular Surgery

**Authors:**
Shuai Wu,[1]† Sophie Leanza,[1]† Lu Lu,[1] Yilong Chang,[1] Qi Li,[1] Diego Stone,[1] Ruike Renee Zhao[1]*

**Affiliations:**
[1]Department of Mechanical Engineering, Stanford University; Stanford, CA 94305, USA.

*Corresponding author. Email: rrzhao@stanford.edu
†These authors contributed equally to this work.

**Abstract:**
Vascular diseases such as thrombosis, atherosclerosis, and aneurysm, which can lead to blockage of blood flow or blood vessel rupture, are common and life-threatening. Conventional minimally invasive treatments utilize catheters, or long tubes, to guide small devices or therapeutic agents to targeted regions for intervention. Unfortunately, catheters suffer from difficult and unreliable navigation in narrow, winding vessels such as those found in the brain. Magnetically actuated untethered robots, which have been extensively explored as an alternative, are promising for navigation in complex vasculatures and vascular disease treatments. Most current robots, however, cannot swim against high flows or are inadequate in treating certain conditions. Here, we introduce a multifunctional and magnetically actuated milli-spinner robot for rapid navigation and performance of various treatments in complicated vasculatures. The milli-spinner, with a unique hollow structure including helical fins and slits for propulsion, generates a distinct flow field upon spinning. The milli-spinner is the fastest-ever untethered magnetic robot for movement in tubular environments, easily achieving speeds of 23 cm·s$^{-1}$, demonstrating promise as an untethered medical device for effective navigation in blood vessels and robotic treatment of numerous vascular diseases.

**One-Sentence Summary:** The magnetic milli-spinner rapidly navigates through blood vessels to deliver drugs and treat aneurysms.

**Main Text:**

# INTRODUCTION

The human vasculature is susceptible to a variety of life-threatening diseases that can manifest all throughout the body (**Fig. 1A**). Among some of the most common diseases are thrombosis (obstruction of blood flow due to clot, **Fig. 1Ai**) (*1*), atherosclerosis (obstruction of blood flow due to plaque buildup, **Fig. 1Aii**) (*2*), and aneurysm (bulging of vessel wall, **Fig. 1Aiii**). Clots and plaque buildup can block blood flow to the heart or brain, which can lead to heart attack or stroke (*3, 4*), while aneurysms in the aorta or cerebral artery can rupture (*5*), causing internal bleeding or stroke. These conditions can be fatal or result in severe damage if not treated in a timely manner, necessitating rapid and reliable treatment methods. State-of-the-art treatment of endovascular disease relies on interventional radiology, a minimally invasive practice that leverages X-ray or other imaging modalities to navigate guidewires and catheters to the occlusion or lesion, after which therapeutic agents or small devices are employed. While generally safer than alternatives such as open surgery, it is difficult to reliably navigate catheters through highly tortuous regions (such as the cerebral arteries in the brain) as they cannot always follow the guidewire, or they require considerable effort to be passed through the vessel (*6*). In addition, vessel perforation/dissection is a concern for regions with especially sharp turns (*7*) or in cases where the catheter-to-vessel diameter ratio is large (*8*). To address the navigation challenges of conventional treatment methods, one approach has been the development of actively controlled guidewires and catheters. When embedded with magnetic-responsive components, these devices become steerable under external magnetic fields, providing enhanced control and maneuverability (*9, 10*), especially at their distal end. However, these magnetically guided tethered devices are still difficult to advance in highly tortuous anatomy due to the low force transmission between the proximal end—at the incision location where the force is applied—and the distal tip of the catheter.

To fundamentally address the navigation difficulties of guidewires and catheters, wireless/untethered magnetic robots have also been extensively explored for navigation in vascular environments and the treatment of vascular diseases. However, without tethers, the controlled navigation of these robots becomes challenging due to the high and pulsatile flow rate of blood (peak velocity of 50-100 cm·s$^{-1}$) (*11*). Several studies tried to overcome this problem by designing robots that generate large friction with the vessel walls to move down or upstream (*12-14*), but they in the meantime pose a high potential to damage the endothelium and can trigger spasms that could block blood flow (*15*). Considering the limitations of recent advances, new untethered robotics capable of complex navigation and rapid, reliable minimally invasive procedures would be important and highly beneficial to enhance the feasibility of future wireless robotic endovascular surgeries.

In this work, we develop a rationally designed multifunctional magnetic millimeter-scale spinner (milli-spinner) robot for the fastest-ever swimming performance in a tubular environment, navigation through the vasculature, and treatment of vascular diseases. These capabilities are demonstrated in *in vitro* flow models of the human pulmonary and cerebral arteries. The milli-spinner is capable of navigating in both straight and tortuous vessel anatomies (**Fig. 1B**), easily overcoming a pulsatile flow with a peak rate of 30 cm·s$^{-1}$ (average rate of 10 cm·s$^{-1}$ or 1 mL·s$^{-1}$), along with the potential to treat diseases. These functionalities of the milli-spinner are enabled by an unprecedented structural design, which induces a unique flow field to facilitate effective endovascular navigation. Featuring a hollow cylindrical body with helical fins, slits, and magnets for actuation (**Fig. 1C**), the milli-spinner generates linear thrust and propulsion upon spinning while allowing fluid to pass through, achieving locomotion without blocking blood flow. A fabricated milli-spinner is shown in **Fig. 1D**, with its size comparison to an aneurysm flow diverter, which is used to reduce blood flow into and occlude an aneurysm after deployment at the lesion (*16*). The milli-spinner can be scaled up or down based on the vessel size and application. Computational fluid dynamics (CFD) simulations and micro-particle image velocimetry (micro-PIV) experiments are performed to understand the influence of the milli-spinner geometry on its performance. Under fluoroscopic guidance (**Fig. 1E**), the milli-spinner can navigate to specific vessel regions under the control of an external rotating magnetic field. Once navigated to the target region, various treatments can be employed including delivery of drugs to targeted regions (**Fig. 1Fi**) to supply localized medicine or perform selective embolization (*in-situ* clotting) of lesions such as aneurysms (**Fig. 1Fii**). Due to its structural design and navigation abilities, the milli-spinner is promising for the effective treatment of numerous vascular diseases throughout the body.

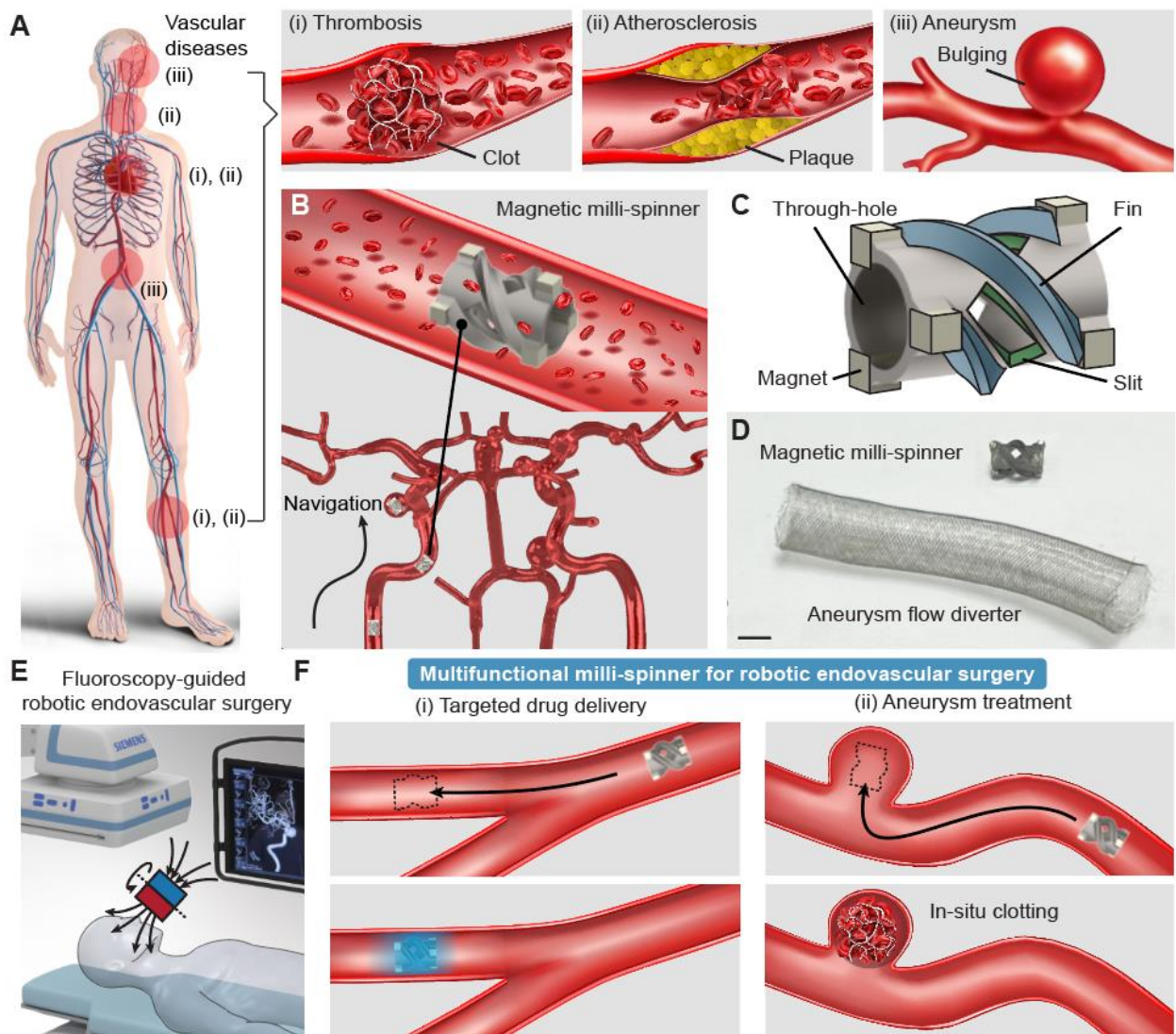

**Fig. 1. Multifunctional untethered magnetic milli-spinner robot for minimally invasive surgeries.** (**A**) Vascular diseases including (i) thrombosis, (ii) atherosclerosis, and (iii) aneurysm. (**B**) Schematic of the developed magnetic milli-spinner navigating in a blood vessel and tortuous endovascular environment. (**C**) Schematic showing the general design of the magnetic milli-spinner. (**D**) Fabricated magnetic milli-spinner compared to an aneurysm flow diverter. Scale bar: 2 mm. (**E**) Schematic of magnetic actuation for fluoroscopy-guided endovascular surgeries by the milli-spinner. (**F**) Multifunctionality of the magnetic milli-spinner for (i) targeted drug delivery and (ii) aneurysm treatment.

## RESULTS

*Magnetic milli-spinner swimming and navigation*

As illustrated in **Fig. 2Ai**, the untethered magnetic milli-spinner is designed from a cylindrical structure with key structural features of a through-hole (dashed yellow line), three helical fins (highlighted in blue), and helical slits (green area). These features play a crucial role in regulating the flow around the milli-spinner, thereby enhancing its swimming performance and enabling other functions. Here, a milli-spinner with a 2.5 mm outer diameter (OD) in a 3.5 mm inner diameter (ID) tube is illustrated as an example. The spinning direction and corresponding moving direction are denoted by the black and blue arrows. We conduct CFD simulations (**Fig. 2Aii**) and micro-PIV (**Fig. 2B**) to show the working mechanism for ultrafast swimming of the milli-spinner. To get visualizable and clean micro-PIV data, we use a relatively low spinning frequency of 2k rpm. From the CFD simulation at the same spinning frequency, the milli-spinner shows a moving speed of 2.6 cm·s$^{-1}$, agreeing well with the micro-PIV measured speed of 2.2 cm·s$^{-1}$. Note that the streamlines in the milli-spinner shown in **Fig. 2Aii** depict an important flow field induced by the unique through-hole and slits features of the milli-spinner, where the flow moves into the hole from the front and either moves out the other end or spins out of the slits. The suction flow entering the hole results from a negative pressure (**Fig. 2Aiii**) in the cavity of the milli-spinner and thus a suction force, and the suction flow is also observed by micro-PIV at the front of the milli-spinner with a maximum speed of 6.0 cm·s$^{-1}$ at the 2k rpm spinning frequency.

To have a better understanding of milli-spinner performance, the swimming speed of the 2.5 mm OD milli-spinner in a 3.5 mm ID tube is assessed by both CFD simulations and experimental measurements at various spinning frequencies, from 1.2k to 12k rpm, as shown in **Fig. 2C**. The rotating magnetic field is homogeneous, provided by Helmholtz coils. The swimming speed increases linearly with the spinning frequency, with a good agreement between the simulation and experiments. Note that the upper limit of the swimming speed for the 2.5 mm OD milli-spinner is not reported here due to the maximum spinning frequency of the magnetic field that can be provided by the Helmholtz coils used. When using higher spinning frequencies in the CFD simulation, for instance 30k rpm, the milli-spinner can reach a swimming speed of 58.6 cm·s$^{-1}$. Meanwhile, tuning the milli-spinner size and design allows for further performance optimization. For example, a 3.5 mm OD milli-spinner can achieve a 56.4 cm·s$^{-1}$ swimming speed at 9.6k rpm spinning frequency. Experimentally, at 8.4k rpm the 2.5 mm OD milli-spinner can already achieve a swimming speed of 14.5 cm·s$^{-1}$, which is sufficient to swim against a pulsatile flow with a peak flow velocity of 30 cm·s$^{-1}$ (average flow rate of 10 cm·s$^{-1}$ or 1 mL·s$^{-1}$), as shown in **Fig. 2D** Note that in the human body, the average flow rate is ~10-20 cm·s$^{-1}$ in veins of the leg (*17*) and ~20-30 cm·s$^{-1}$ in the internal carotid artery (ICA, ~4-6 mL·s$^{-1}$, peak flow velocity of ~60 cm·s$^{-1}$) (*18*).

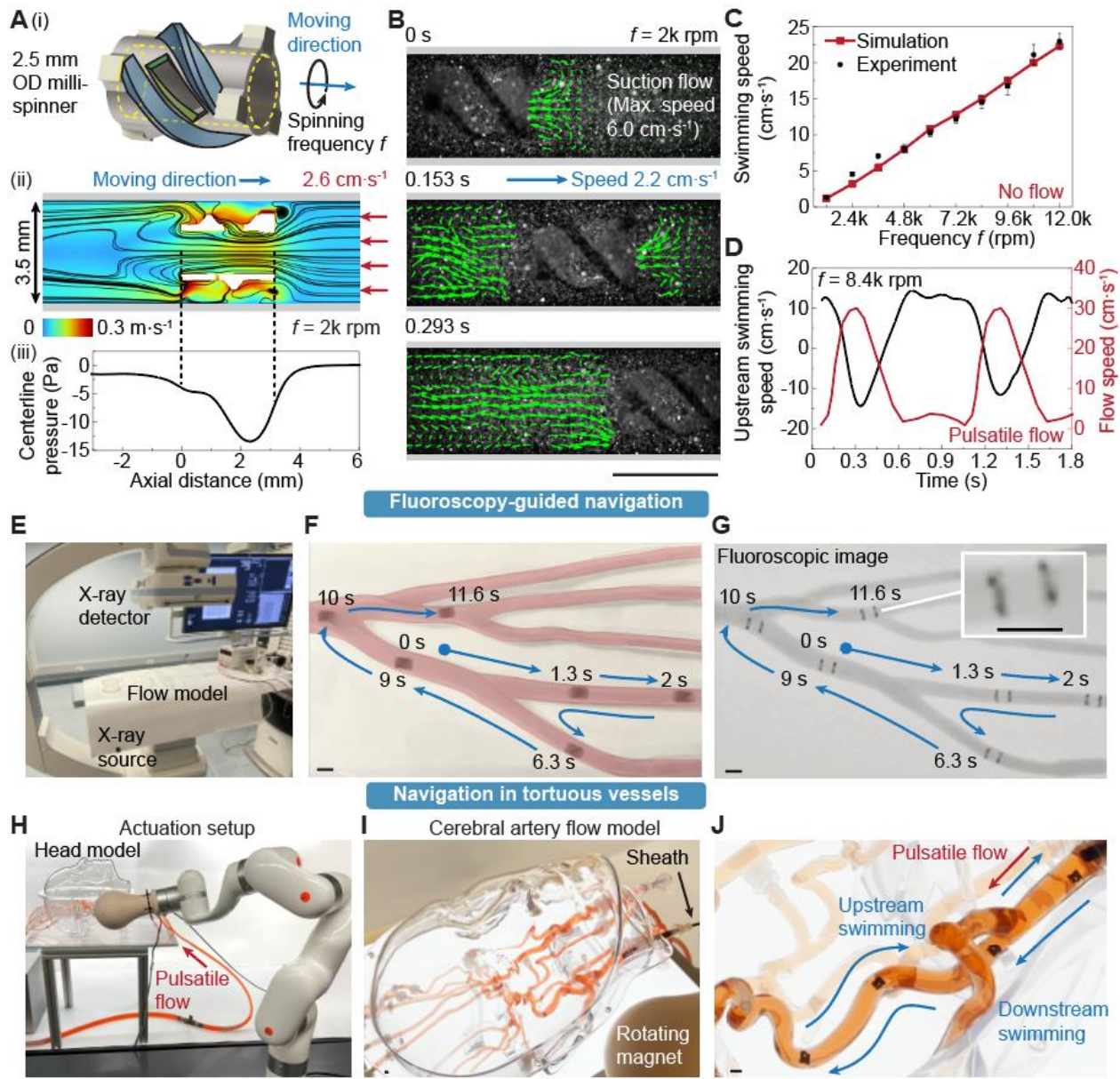

**Fig. 2. Magnetic milli-spinner swimming and navigation.** (**A**) (i) A 2.5 mm OD milli-spinner with fins, a through-hole, and slits. (ii) CFD simulation and streamlines for the milli-spinner at 2k rpm spinning frequency in a 3.5 mm ID tube. The hole and slit features induce a flow field in which flow moves into the hole and either moves out the other end or spins out of the slits. The flow entering the hole results from (iii) the negative pressure generated within the milli-spinner. (**B**) PIV of the milli-spinner swimming at 2k rpm in a 3.5 mm ID tube. (**C**) Milli-spinner swimming speed characterization at varied spinning frequencies in a 3.5 mm ID tube. (**D**) Milli-spinner swimming at 8.4k rpm against a pulsatile flow with 30 cm·s$^{-1}$ peak velocity and 60 beats per minute. (**E**) Setup for fluoroscopy-guided navigation in a pulmonary artery flow model. (**F**) Photo and (**G**) fluoroscopic image showing the real-time navigation of the milli-spinner through the pulmonary artery flow model. The zoomed-in image shows the milli-spinner, which is visible due to the attached magnets at both of its ends. (**H**) Control of the milli-spinner by a robotic arm. (**I**) Experimental setup of a cerebral artery flow model. (**J**) Milli-spinner navigating in a tortuous vessel with and against flow. Scale bars: 3.5 mm.

While the milli-spinner is highly effective in a straight tube (**Fig. 2A** – **Fig. 2D**), it is important to assess the milli-spinner's performance in clinically-relevant vasculatures. Here, utilizing the agile motion control enabled by moving a rotating magnet, the milli-spinner is navigated through a pulmonary artery flow model under real-time fluoroscopy guidance (**Fig. 2E**). This flow model offers a relatively simple environment through which the milli-spinner's steering capabilities can be evaluated. When encountering a branching point, the milli-spinner can navigate into either branch as determined by the orientation of the magnetic field's rotational axis. As illustrated in **Fig. 2F**, the magnetic milli-spinner approaches the bottom branching point and navigates to its top branch in 2 s. The milli-spinner can return along the same path by simply reversing the rotational direction of the applied magnetic field without requiring a 180-degree turn in such a confined space. Lastly, the milli-spinner is guided to reach the bottom branch at 6.3 s and eventually reaches the top branching point at 11.6 s. The magnets (with density around 7.6 g·cm$^{-3}$) allow for tracking of the milli-spinner due to their good visibility under X-ray imaging, even when being obstructed by a skull bone (density of 1.6 to 1.9 g·cm$^{-3}$) (*19*). As shown in **Fig. 2G**, the milli-spinner location is evident from the magnets, which are the dark dots at the milli-spinner's two ends under X-ray.

The navigation of more complex vasculature, such as the highly tortuous three-dimensional (3D) cerebral arteries, poses significant challenges. Such vasculature requires precise and dynamic control of the milli-spinner's motion. For these scenarios, a six-axis robotic arm can provide multi-directional control (**Fig. 2H**), enhancing the precision of the milli-spinner's movement. This robotic system enables navigation through the intricate and convoluted pathways characteristic of cerebral arteries (**Fig. 2I**) by adjusting the milli-spinner's orientation and trajectory in real-time. Combining the milli-spinner's inherent navigation capabilities with the advanced control provided by the robotic arm, the precision required for complex endovascular procedures is achievable. As shown in **Fig. 2J**, the milli-spinner is first delivered to the cerebral arteries through a sheath and swims downstream to the tortuous vessel in a pulsatile flow of 1 mL·s$^{-1}$ (average flow rate of 5 to 8 cm·s$^{-}$ along the path) at 5.4k rpm spinning frequency. While returning, the milli-spinner can be controlled to swim upstream under a higher spinning frequency of 7.2k rpm to overcome the flow. As the milli-spinner gets close to the sheath, aspiration can be applied to suck the milli-spinner back into the sheath. This approach could potentially improve the safety of traditional minimally invasive interventions in highly tortuous vasculature.

*Magnetic milli-spinner for targeted drug delivery*

In this section, we explore the rotational modes of the milli-spinner and how switching its rotation axis can be used to control drug release rates for targeted drug delivery. Based on the frequency and the magnitude of the rotating magnetic field, there are two distinct motion modes that the milli-spinner can achieve, namely spinning and flipping (**Fig. 3A**), with the magnetic field rotating about the dashed black line in the schematic. The milli-spinner undergoes a spinning motion when it rotates about its longitudinal axis (axis of the milli-spinner's cylindrical structure). On the other hand, for the flipping motion, the magnetic field's rotational axis is perpendicular to the milli-spinner's longitudinal axis. As shown in the contour plot of **Fig. 3B**, the two motion modes, spinning and flipping, can switch between each other by adjusting the magnetic field magnitude and frequency. For instance, at 15 mT, the milli-spinner tends to flip when the rotational magnetic field frequency is relatively low (below ~3.6k rpm). As the frequency increases to 4.8k rpm, the milli-spinner transitions into the spinning mode. However, if the frequency becomes too high (above ~7.2k rpm), the milli-spinner may become unstable and unable to follow the rotational magnetic field. To achieve stable spinning at higher frequencies, stronger magnetic fields are required (e.g., 20 mT for 8.4k rpm spinning).

The different motion modes of the milli-spinner can be exploited for functionalities such as controlled drug delivery. As shown in **Fig. 3C**, the magnetic milli-spinner's hollow structure allows for drug storage, with the drug sealed inside by soluble covers. The milli-spinner can then release the drug at different rates depending on its motion mode under varying magnetic field conditions.

A proof-of-concept demonstration for controlled drug release using the milli-spinner is shown in **Fig. 3D** and **Fig. 3E**, in which blue powder is loaded and sealed into the milli-spinner as a drug model. With a spinning motion, the milli-spinner gradually releases the drug as it moves back and forth, and the blue color slowly intensifies over time (**Fig. 3D**). Alternatively, when the milli-spinner flips, the loaded drug is released rapidly in just 5-10 s, as evidenced by the sudden burst of blue color (**Fig. 3E**). The magnetic milli-spinner's targeted drug release capability is also demonstrated in a 1:1 cerebral artery flow model (**Fig. 3F**). Dense powders (copper, 40 µm) are loaded and sealed into the milli-spinner to illustrate the fluoroscopy-guided targeted drug release process. Initially, the milli-spinner body is visible due to the dense powder in its cavity, as illustrated in **Fig. 3F**. Following a programmed navigation path, the milli-spinner reaches the targeted middle cerebral artery (MCA) in 9 s, after which the seals gradually dissolve and release the powder. The drug release process is also indicated by the progressive fading of loaded powder in the milli-spinner cavity in the fluoroscopic images, with only the magnets remaining visible once the drug has been fully dispensed (**Fig. 3F**).

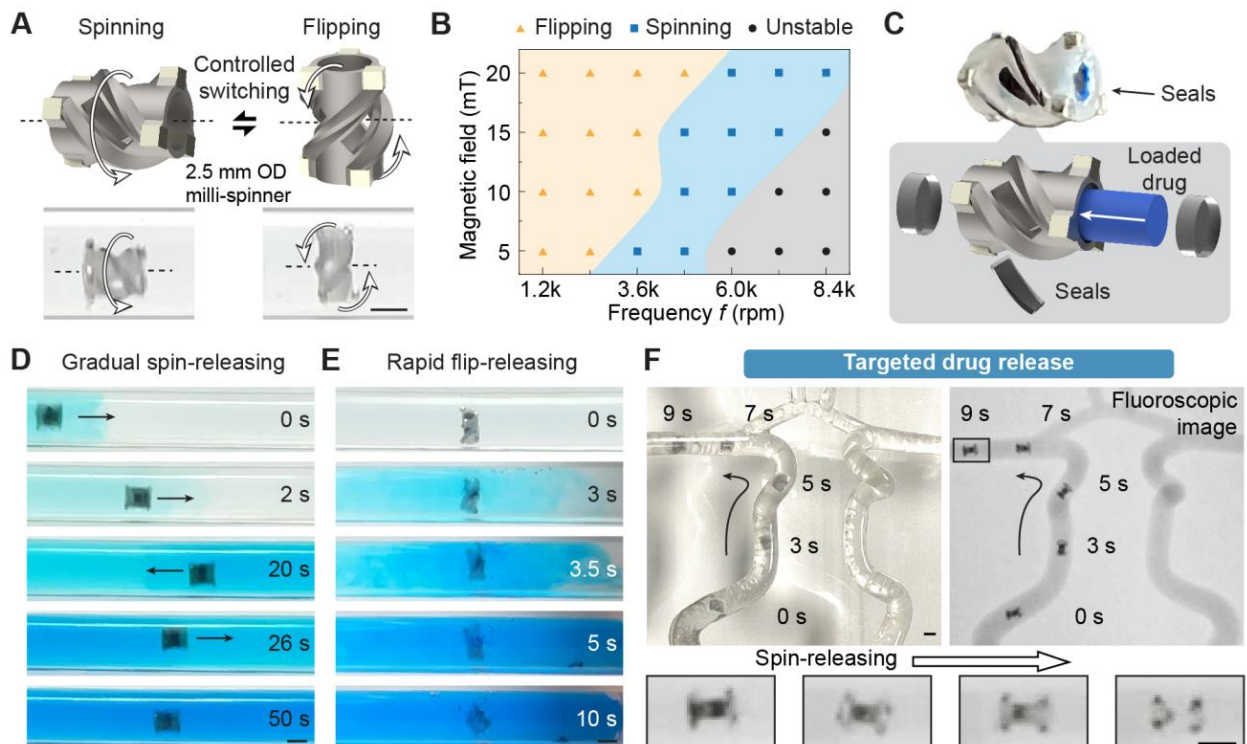

**Fig. 3. Controlled magnetic milli-spinner motion for drug release.** (**A**) Two distinct motion modes of the milli-spinner: spinning and flipping, during which the milli-spinner rotates about the same axis as the rotational magnetic field, indicated by the dashed black line. (**B**) Contour plot illustrating milli-spinner motion modes under various combinations of magnetic field magnitudes and frequencies. (**C**) Schematics and fabricated sample of the milli-spinner loaded with a model drug. (**D**) Snapshots of gradual drug release facilitated by the spinning motion. (**E**) Snapshots of rapid drug release achieved through flipping motion. (**F**) Targeted drug release under fluoroscopic guidance. Scale bars: 2.5 mm.

*Magnetic milli-spinner for aneurysm treatment*

Aneurysms are weakened, bulging areas of artery walls that pose a significant risk of rupture, leading to severe medical complications, for example, hemorrhagic stroke if occurring in the cerebral artery (*20*). Current minimally invasive surgeries for aneurysms rely on interventional procedures that use guidewires and catheters to navigate to the lesion and perform treatment. These procedures, such as aneurysm flow diverter placement or endovascular coiling, aim to achieve selective embolization (in-situ clotting) and reduce/prevent blood flow into the weakened area (*6*). However, current treatment methods can be challenging due to difficulties with catheter/guidewire navigation in complex and tortuous blood vessel anatomy.

In this study, we demonstrate an innovative approach to aneurysm treatment using a magnetic milli-spinner under fluoroscopic imaging guidance (**Fig. 4A**). The procedure begins with the identification of the aneurysm location via digitally subtracted angiography, a technique that uses X-ray to obtain the blood vessel map by injecting contrast dye. Once the aneurysm is identified, the milli-spinner is introduced through a sheath and swims through the tortuous and distal vessels to reach the aneurysm site for embolization. Two different mechanisms for in-situ embolization are explored. The first mechanism uses the milli-spinner to introduce embolic agents (coagulation agents illustrated in **Fig. 4B**) to the targeted lesion. In this demonstration, the cerebral artery flow model is filled with anticoagulant-treated porcine blood (**Fig. 4C**). After injecting contrast dye, the pre-treatment angiogram shows a sac structure connected to the regular vessel with a profile denoted by the dashed black line, indicating the aneurysm location, as shown in **Fig. 4D**. Under fluoroscopic guidance, the milli-spinner then navigates into the target aneurysm, where its seals dissolve, releasing the coagulation agent through its spinning motion (**Fig. 4E**). Calcium chloride is used as the coagulation agent to counteract the anticoagulant in the blood, triggering clot formation around the magnetic milli-spinner within the model aneurysm. Post-treatment imaging (**Fig. 4F**) shows that the aneurysm is successfully filled, with no contrast dye entering the aneurysm denoted by the dashed black line, confirming effective embolization. Although the milli-spinner remains within the aneurysm, it would not cause safety concerns after treatment due to its small magnets and weak magnetic force exerted on the clot and surrounding vasculature even when exposed to strong external magnetic fields. The second mechanism leverages the milli-spinner's ability to deliver expandable polymeric materials for aneurysm treatment (**Fig. 4G**). As shown in **Fig. 4H**, the milli-spinner carrying an expandable material navigates to the aneurysm. After reaching the target position, the expandable material gradually absorbs liquid, significantly increasing in volume (white dashed line) to fill the aneurysm (black dashed line) as shown in **Fig. 4I**. After the material has fully expanded with the milli-spinner remaining in the aneurysm, a blue-colored fluid is flown through (**Fig. 4J**), where it is seen that the blue fluid does not mix with the red material in the aneurysm, indicating reduced flow into the aneurysm. The two presented embolization techniques can serve as powerful and quick methods for targeted aneurysm treatment in tortuous vasculature.

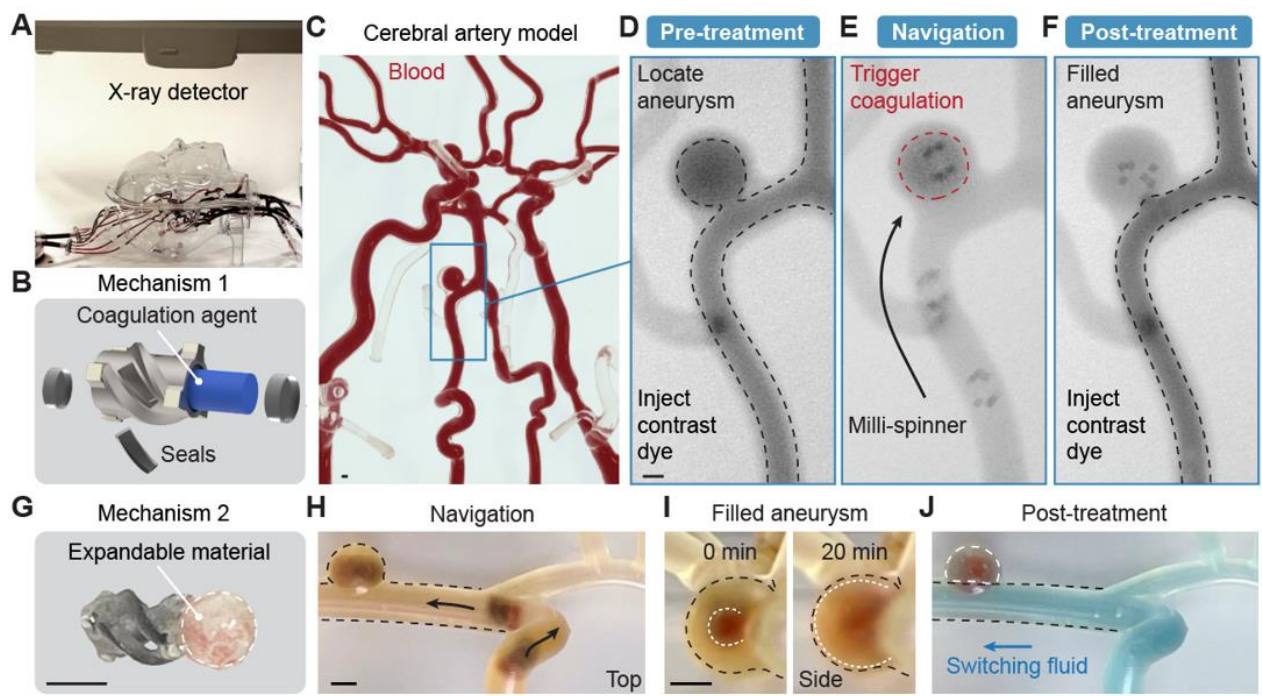

**Fig. 4. Magnetic milli-spinner for aneurysm treatment via in-situ embolization.** (**A**) Experimental setup with fluoroscopic guidance. (**B**) Mechanism 1: milli-spinner loaded with coagulation agent for in-situ embolization. (**C**) Cerebral artery flow model filled with anticoagulant-treated porcine blood. (**D**) Pre-treatment angiogram shows the location of the aneurysm. (**E**) Milli-spinner navigates to the aneurysm and spin-releases drug, triggering coagulation in the aneurysm. (**F**) Post-treatment angiogram demonstrates the filled aneurysm due to in-situ embolization by the milli-spinner. (**G**) Mechanism 2: milli-spinner attached with expandable material for in-situ embolization. (**H**) Milli-spinner with attached expandable material navigates to aneurysm location. (**I**) Expandable material attached to the milli-spinner absorbs water and fills the aneurysm. (**J**) Post-treatment images showing blocking of blue fluid into the filled aneurysm. Scale bars: 2.5 mm.

## DISCUSSION

This study presents a magnetic milli-spinner, the fastest-swimming endovascular robot to date, capable of agile navigation in complex vasculature and overcoming high-flow conditions, enabling a range of robotic endovascular interventions of targeted drug delivery and aneurysm treatment. The unique hollow structure of the milli-spinner with helical fins and slits allows for propulsion upon spinning, resulting in rapid swimming under remote actuation without blocking blood flow. The milli-spinner can serve as a vessel for targeted drug delivery, accommodating various drug release rates through the control of its motion modes—either spinning or flipping. Meanwhile, the milli-spinner shows promise for effective aneurysm treatment in complex and tortuous blood vessel anatomies that are difficult for guidewires and catheters to navigate through.

While the magnetic milli-spinner is promising for navigation in complex vasculature and endovascular surgeries, there are a few aspects which should be further studied to improve its abilities. Firstly, CFD simulations can be taken advantage of for design optimization of the milli-spinner. As there is an extensive design space including numerous geometric features of the milli-spinner as well as different flow conditions and vascular geometries, the milli-spinner can be optimized for better performance in different environments. Secondly, in highly tortuous 3D vessels, the robot requires more precise control strategies. Currently, X-ray imaging informs on the position of the milli-spinner, which dictates where the rotating magnet is moved to in *in vitro* flow models. However, in reality, real-time imaging should be used for closed-loop control of the robotic arm and its rotating magnet to keep track of the milli-spinner's position and guide its orientation for future movements. Accordingly, algorithms can be developed which utilize 2D X-ray images and map the position of the milli-spinner to its location within the 3D vasculature, informing on the necessary position of the rotating magnet for the milli-spinner to reach a desired location or traverse a specific path. Once this is achieved, the milli-spinner would be able to operate autonomously, which could relieve the burden on interventional radiologists and ensure more reliable medical procedures within tortuous vasculature. We envision that such advancements made to our current technology would lead to more widespread availability of robotic endovascular treatments, improving outcomes for patients facing a variety of endovascular conditions.

## MATERIALS AND METHODS

*Magnetic milli-spinner fabrication.*

Magnetic milli-spinners are composed of two components: the main body and attached neodymium iron boron magnets. The milli-spinner body was 3D printed using a customized digital light processing printer with a printing resolution of 9 µm. The printer includes a 385 nm UV light projector module (PRO4500, Wintech Digital Systems Technology Corp., USA), a 4× lens (Nikon Corp., Japan), a resin tank with a Teflon AF window (70 µm thickness, VICI Metronics, Inc., USA), and a translation stage (LTS150/M, Thorlabs, USA). The printing resin was prepared with 99.7 wt% Formlabs Tough 2000 (Formlabs Inc., USA) and 0.3 wt% iron oxide (300 nm, Alpha Chemicals.,

USA), which was mixed at 2000 rpm for 30 s (AR-100, Thinky, USA) to ensure material homogenization. Printing parameters of 2.36 mW·cm$^{-2}$ light intensity, 25 µm layer thickness, and 2.5 s layer curing time are used. Once post-processing the printed main body, three cube magnets (N50 neodymium, SM Magnetics, USA) were glued to each side for the 2.5 mm OD milli-spinners in **Fig. 2**, **Fig. 3**, and **Fig. 4**.

*Magnetic actuation setup.*

Both electromagnetic coils and magnets can provide the magnetic field required to actuate the magnetic milli-spinner. Different setups of electromagnetic coils or motor-driven magnets were adopted for different demonstrations.

*Milli-spinner for drug delivery.*

In **Fig. 3D-F**, food dye or copper particles (40 µm, EnvironMolds, USA), serving as drug mimics, were loaded into the milli-spinner cavity, followed by sealing of the milli-spinner with a water-soluble material (Hypromellose). The filled milli-spinner was then placed in an oven at 80°C for one hour to secure the seal.

*In-situ clotting and aneurysm treatment.*

For the milli-spinner aneurysm treatment mechanism 1 demonstration in **Fig. 4C**, anticoagulated porcine whole blood with sodium citrate (3.8 wt%) in the volume ratio of 9:1 was obtained from Animal Technologies, Inc. Calcium chloride (Aldon Corp., USA), which reversed the effect of the anticoagulant, served as a coagulation agent and was sealed in the milli-spinner via a water-soluble material.

For the milli-spinner aneurysm treatment mechanism 2 demonstration in **Fig. 4H**, the expandable material (Polyacrylate) was attached to the milli-spinner and coated with a water-soluble layer that delays the material's expansion.

**Acknowledgments:**
The authors acknowledge the support of NSF Career Award CMMI-2145601, National Institutes of Health Shared Instrumentation grant S10RR026714. The authors would like to acknowledge Robert Bennett for assisting in the fluoroscopic imaging experiments, Professor Juan G. Santiago for the support on the Micro-PIV setup and the valuable insights on Micro-PIV results, and Six Oliva Skov for helping build Micro-PIV setup.

**Author contributions:**
    Conceptualization: RRZ
    Methodology: SW, RRZ
    Investigation: SW, SL, LL, YC, QL, DS
    Visualization: SW
    Funding acquisition: RRZ
    Project administration: RRZ
    Supervision: RRZ
    Writing – original draft: SW, SL, RRZ
    Writing – review & editing: SW, SL, LL, YC, RRZ

**Competing interests:** One PCT application has been filed on the reported technology.

**Data and materials availability:** The authors declare that the data that support the findings of this study are available from the corresponding author upon the reasonable statement.